\documentclass[conference]{IEEEtran}
\IEEEoverridecommandlockouts
\usepackage{cite}
\usepackage{amsmath,amssymb,amsfonts}
\usepackage{algorithmic}
\usepackage{graphicx}
\usepackage{textcomp}
\usepackage{xcolor}
\usepackage{epsfig}
\usepackage{float}
\usepackage{caption}
\usepackage{subcaption}
\usepackage{eucal}

\def\BibTeX{\rm B\kern-.05em{\sc i\kern-.025em b}\kern-.08em T\kern-.1667em\lower.7ex\hbox{E}\kern-.125emX}
\begin{document}

\title{Dynamic Bayesian Approach for decision-making in Ego-Things}

\author{\IEEEauthorblockN{Divya Kanapram\textsuperscript{1,2},Damian Campo\textsuperscript{1}, Mohamad Baydoun\textsuperscript{1}, Lucio Marcenaro\textsuperscript{1},\\ Eliane L. Bodanese\textsuperscript{2},
Carlo Regazzoni\textsuperscript{1}, Mario Marchese\textsuperscript{1}}
\IEEEauthorblockA{\textit{Department of Electrical, Electronics and Telecommunication Engineering and Naval Architecture,} \\ \textit{University of Genova, Italy}\textsuperscript{1}\\
\textit{School of Electronic Engineering and Computer Science (EECS),}\\ 
\textit{Queen Mary University of London, UK}\textsuperscript {2}} 
}

\IEEEoverridecommandlockouts
\IEEEpubid{\makebox[\columnwidth]{978-1-5386-4980-0/19/\$31.00 \copyright2019 IEEE} \hspace{\columnsep}\makebox[\columnwidth]{ }}

\maketitle

\begin{abstract}
This paper presents a novel approach to detect abnormalities in dynamic systems based on multisensory data and feature selection. The proposed method produces multiple inference models by considering several features of the observed data. This work facilitates the obtainment of the most precise features for predicting future instances and detecting abnormalities. Growing neural gas (GNG) is employed for clustering multisensory data into a set of nodes that provide a semantic interpretation of data and define local linear models for prediction purposes. Our method uses a Markov Jump particle filter (MJPF) for state estimation and abnormality detection. The proposed method can be used for selecting the optimal set features to be shared in networking operations such that state prediction, decision-making, and abnormality detection processes are favored. This work is evaluated by using a real dataset consisting of a moving vehicle performing some tasks in a controlled environment.
\end{abstract}


\section{Introduction}
Internet of Things (IoT) is a paradigm that is rapidly gaining attention in the field of modern information and communication technologies (ICT). The basic idea of this concept is to establish wired or wireless connectivity among a variety of things and enable addressing schemes to interact and cooperate with external entities to create new applications/services and achieve common goals \cite{atzori2010internet}. Many challenging issues still need to be solved in IoT; among them, some challenges involve full operability of interconnected things, a high degree of smartness in connected devices, guaranteeing trust, privacy and security. In this paper, we mainly focus on developing smartness in things that are part of IoT and investigate how different features of a single \textit{thing} can contribute to achieving an objective. Even with the current developments in artificial intelligence and machine learning, we still lack a genuine theory that describes the underlying principles and methods that allow designed things not only to understand their environment but also be conscious about what they do. Such artificial consciousness would lead to a new generation of systems (things) that take timely initiatives beyond the already programmed goals set by humans. This work focuses on developing smartness in things by exploiting different features in observed data. Moreover, we investigate the possibility to build an autonomous decision system based on feature selection \cite{rao2018feature} to take appropriate actions in different situations. 

IoT requires to be smarter to do more than sensing and sometimes actuating \cite{green2015internet}. Another similar argument is that ``without comprehensive cognitive capability, IoT is just like an awkward stegosaurus: all brawn and no brains'' \cite{wu2014cognitive}. Things in IoT face new challenges in understanding their surroundings,i.e., \textit{situational-awareness}; and their own states, i.e., \textit{self-awareness}. Both types of awareness are fundamental to perform efficient decision making and cooperate with other external agents. 

The concept of self-awareness is not new; it has been proposed for Autonomic computing more than a decade ago. In the seminal 2001 paper \cite{1160055}, Kephart and Chess envision ``computing systems that can manage themselves given high-level objectives from administrators.'' Another hypothesis has been mentioned in \cite {chatila2017toward}, which states that self-awareness must first rely on the perception of self as different from the environment and other agents. To achieve this, firstly the entity needs to interact with the environment and build sensory-motor representation; then needs to interact with other entities to distinguish it from them. In this paper, we develop different models from the data captured by various sensors and make performance comparisons, i.e., which model could be best suited to predict the future states of the entity in different situations. 

This work is highly motivated by models proposed in \cite{baydoun2018multi, baydoun2018learning}. In \cite{baydoun2018multi}, the authors propose a method to develop a multi-layered self-awareness in autonomous entities and exploit this feature to detect abnormal situations in a given context. On the other hand, \cite{baydoun2018learning} introduces a Markov Jump Particle filter (MJPF), which consists of continuous and discrete inference levels that are dynamically estimated by a collection of Kalman filters (KF) assembled into a Particle filter (PF) algorithm.MJPF enables the prediction of future states in continuous and discrete levels. Most of the related works \cite{ravanbakhsh2018learning,han2017tdoa} use position-related information to make inferences.However, the information related to the control of an autonomous entity plays a significant role in the prediction of future states and actions of the entity. By taking this into account, it is imperative to consider variables related to control to make the system more efficient.

This work is motivated by the ideas developed in \cite{baydoun2018multi}. Nonetheless, the novelties of this article are listed as follows: \textit{i)} It includes a novel strategy for segmenting and interpreting clusters of generalized states. It uses growing neural gas (GNG) to obtain them. \textit{ii)} It incorporates a probabilistic anomaly measurement based on the Hellinger distance. \textit{iii)} It considers multisensory data containing information of actuators. Observations are divided into different sets of data, generating multiple inference subsystems. \textit{iv)} It proposes a strategy to select the set of sensory data (features) that optimize the prediction of future states and the detection of abnormalities.

The rest of the paper is organized as follows. Section II describes the proposed methodology. Section III presents the experimental results. In section IV, conclusion and future work are highlighted.

\section{methodology}

This section discusses how ``awareness'' capabilities are modeled into things/entities leading to the generation of ``ego-things''. Accordingly, \textit{Ego-Things} are defined as intelligent systems that perceive their internal parameters and adapt themselves when facing non-stationary situations \cite{camara2017self}. Ego-Things are endowed with several sensors that enable a synchronized multi-view perception of their actions.  In this work, time-synchronized sensory data related to control (actuation) part of the ego-thing are analyzed. The observed multisensory data have been categorized into different groups in order to learn a number of DBN models and select the most suited one for enriching efficiency when predicting future states and recognizing anomalies. 

\subsection{Generalized states}
By considering an ego-thing endowed with a certain amount of sensors that monitors its activity, it is possible to define $Z^m_k$ as any combination (indexed by $m$) of the available measurements in a time instant $k$. Accordingly, let $X^m_k$ be the states related to measurements $Z^m_k$, such that: $Z^m_k = X^m_k + \omega_k$; where $\omega_k$ represents the noise associated with the sensor devices. Additionally, let define the generalized states associated with a sensor combination $m$ as:
\begin{equation}\label{eq1}
\boldsymbol{X}_k^m = [X_k^m \hspace{2mm}  \dot{X}_k^m \hspace{2mm} \ddot{X}_k^m  \hspace{0.5mm} \cdots \hspace{0.5mm} X^{m,(L)}_k]^\intercal,
\end{equation}
where $(L)$ indicates the $L$-th time derivative of the state.
\subsection{Learning DBN}

After obtaining the generalized states from strings of observed data, the next step consists of learning a switching DBN model that predicts the evolution of the system's states over time. We learn a DBN for each combination of sensors $m$ such that a set of DBNs is generated and written as:
\begin{equation}\label{eq2}
 \boldsymbol{DBN} = \{DBN_1, \cdots, DBN_m, \cdots, DBN_M\}.
\end{equation}
Each sensor combination $m$ can be seen as a feature used to learn the parameters of a DBN architecture. Consequently, $DBN_m$ represents the DBN learned from the feature $m$. $M$ is the total number of features taking into consideration.  

\begin{figure}[t]
\centering
 	\includegraphics[width = 1 \linewidth ]{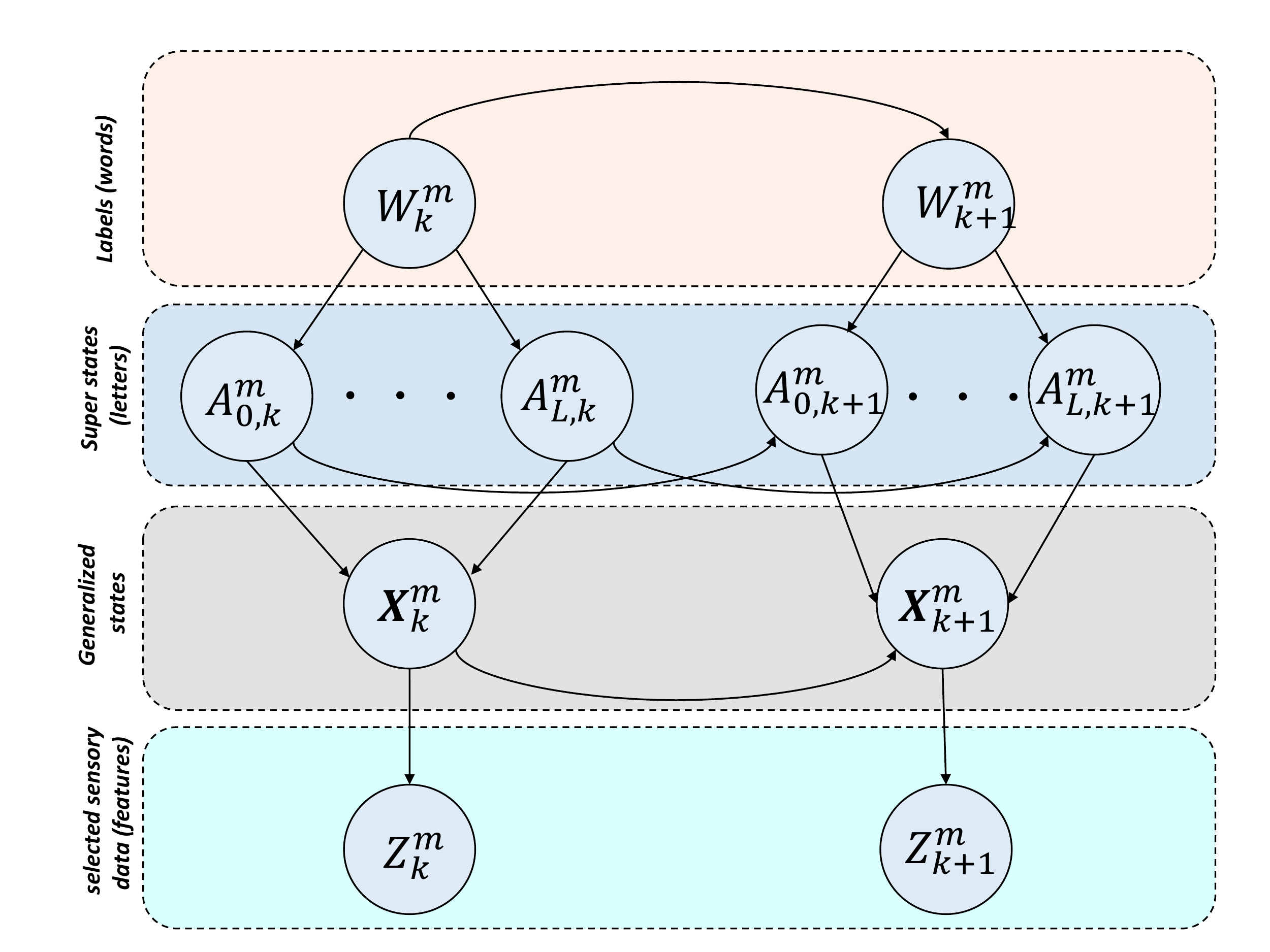}
\caption{Proposed DBN}
\label{fig:DBN}
\end{figure}

The same DBN architecture is considered for making inferences with different proposed features. As can be seen in Fig. \ref{fig:DBN}, the proposed DBN consists of four hierarchical levels. The two lower levels (cyan and grey backgrounds) represent the continuous information of the DBN. Moreover, the other two semantic levels are considered on the top of the architecture (blue and orange backgrounds). Such semantic levels enable a discrete interpretation of data such that a linguistic understanding of sensory data can be achieved by machines. 

For obtaining discrete inference levels, a clustering algorithm based on GNG is used to segment the state space into various regions according to their time derivative order. In GNG, the learning is continuous and the addition/removal of nodes is automated \cite{Fritzke:1994:GNG:2998687.2998765}. These are the reasons to choose GNG algorithm over other clustering algorithms such as K-means or SOM.

A total number of $L + 1$ GNGs are trained for each set of data, where $L$ is the upper limit of the states time derivatives, see equation \ref{eq1}. The output of each GNG consists of a set of nodes that encode constant behaviors for a specific state's time derivative order. In other words, at a time instant, each GNG takes the data related to a single time derivative $X_k^{m,(l)} \in \boldsymbol{X}_k^m $ and cluster it with the same time derivative data acquired in previous time instances. Generated GNG nodes can be considered as \textit{alphabets} and they represent the encoded relevant dynamics of the ego-thing. The set of nodes encoding time derivatives of order $l$-th for the feature $m$ is defined as follows:
\begin{equation}\label{eq3}
A^m_l = \{\bar{X}^{m,(l)}_{1}, \bar{X}^{m,(l)}_{2}, \cdots  \bar{X}^{m,(l)}_{N_l}\},
\end{equation}

where $N_l$ is the number of nodes in the GNG associated to the $l$-th order time derivative. Each node in a GNG represents the centroids of associated data point clusters. 

In each feature-case, by taking into consideration all possible combinations of centroids among the involved GNGs, we obtain a set of \textit{words}, which define generalized states in a semantic way. Such words have the following structure:

\begin{equation}\label{eq4}
W^m = \{\alpha, \dot{\alpha}, \cdots  \alpha^{(L)}\},
\end{equation}

where $\alpha^{(l)} \in A_l$. $W_m$ contains all possible combinations of discrete generalized states and can be seen as a \textit{dictionary} that puts together all possible letters from the learned alphabets. $W$ is a high-level hierarchy variable that explains the generalized states from a semantic viewpoint. The generation of alphabets and dictionaries is executed based on the observed training samples. 

To make inferences by employing the learned DBN, we proposed to use a probabilistic switching model called MJPF \cite{baydoun2018learning}. In this filter, a KF is used to model continuous level information and a PF algorithm employed to estimate discrete level information of the ego-thing. When observing new data, in time instant, MJPF facilitates to make real-time predictions of future generalized states based on the semantics learned through the GNG.

\subsection {Online prediction and abnormality detection}

As an abnormality measurement, it is used the Hellinger distance \cite{Lourenzutti2014} between predicted generalized states and observation evidence. Accordingly, let $p(\boldsymbol{X}^m_k|\boldsymbol{X}^m_{k-1})$ be the predicted generalized states and $p(Z_k|\boldsymbol{X}^m_k)$ be the observation evidence. The Hellinger distance can be written as:
\begin{equation}\label{eq6}
\theta_k^m = \sqrt{1 - \lambda_k^m},
\end{equation}
where $\lambda_k^m$ is defined as the Bhattacharyya coefficient \cite{Bhattacharyya1943}, such that:

\begin{equation}\label{eq7}
\lambda^m_k = \int \sqrt{p(\boldsymbol{X}_k^m|\boldsymbol{X}_{k-1}^m) p(Z_k^m|\boldsymbol{X}_k^m)} \hspace{1mm} \mathrm{d}\boldsymbol{X}_k^m.
\end{equation}

As can be seen in equation \eqref{eq6}, when evaluating a given experience, at each time instant it is obtained an abnormal measurement for each feature $m$. Such consideration makes it possible to compare abnormal behaviors under several sensory assumptions (features) and select the one that precisely predicts future instances and detects abnormal situations. The variable $\theta_k^m \in [0,1]$, where values close to $0$ indicate that measurements match with predictions; whereas values close to $1$ reveal the presence of an abnormality. 

\subsection{model selection}
After learning a set of DBNs, it is possible to select the model that detect abnormalities more accurately. Since each DBN produces a set of abnormality measurements, see equations \eqref{eq6} and \eqref{eq7}, a supervised approach where the testing data is already labelled in abnormal/normal samples; is employed for measuring the performance of the different DBNs when detecting anomalies. Accordingly, the true positive rate ($TPR$) and false positive rate ($FPR$) are obtained making it possible to build a set of ROC curves each of them containing the performance of a given DBNs. As is well known, ROC curves plot $TPR$ and $FPR$ at different thresholds, where:
\begin{equation}\label{eq8}
TPR = \frac{TP}{TP + FN} \hspace{1mm}; \hspace{4mm} FPR = \frac{FP}{FP + TN}
\end{equation}
In the proposed context, $TP$ is defined as the number of times where abnormalities are correctly identified. $FN$ consists of the times that abnormalities are classified incorrectly. Accordingly,  $FP$ are the times where anomalies are wrongly assigned to normal samples and $TN$ represents the times where normal samples are correctly identified .  

This work considers two different measurements for selecting the most precise DBN. They are: \textit{(i)} the area under the curve ($AUC$) of the ROC curves, which quantifies the performance of the DBNs' abnormal detection at several thresholds. \textit{(ii)} The accuracy ($ACC$) measurement, which is defined as follows: 
\begin{equation}\label{eq9}
ACC = \frac{TP + TN}{TP +TN +FP + FN},
\end{equation}
In our context, for each DBN both measurements are obtained and compared for selecting the most precise DBN for abnormality detection purposes. 

\section{EXPERIMENTAL RESULTS}
\subsection{Dataset}
\begin{figure}[h]
	\begin{subfigure}[t]{0.25\textwidth}
		\centering
		\includegraphics[width=3.75cm, height=3cm]{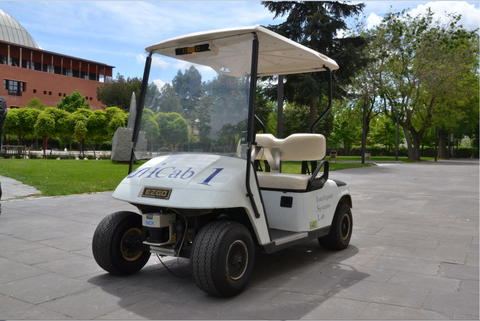}
		\caption{The autonomous vehicle “iCab”}
		\label{fig:iCab}
	\end{subfigure}%
	\begin{subfigure}[t]{0.25\textwidth}
		\centering
		\includegraphics[width=3.5cm, height=3cm]{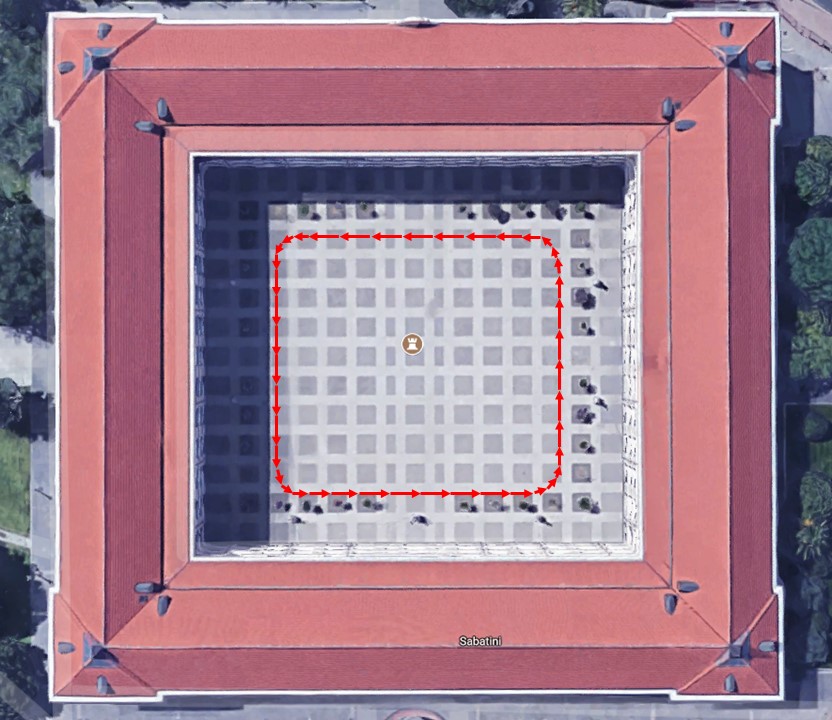}
		\caption{The environment}
		\label{fig:environment}
	\end{subfigure}
	\caption{The agent and the environment used for the experiments.}
	\label{fig:icabAndEnv}
\end{figure}

\begin{figure}[h]
	\begin{subfigure}[t]{0.25\textwidth}
		\centering
		\includegraphics[width=0.85\linewidth]{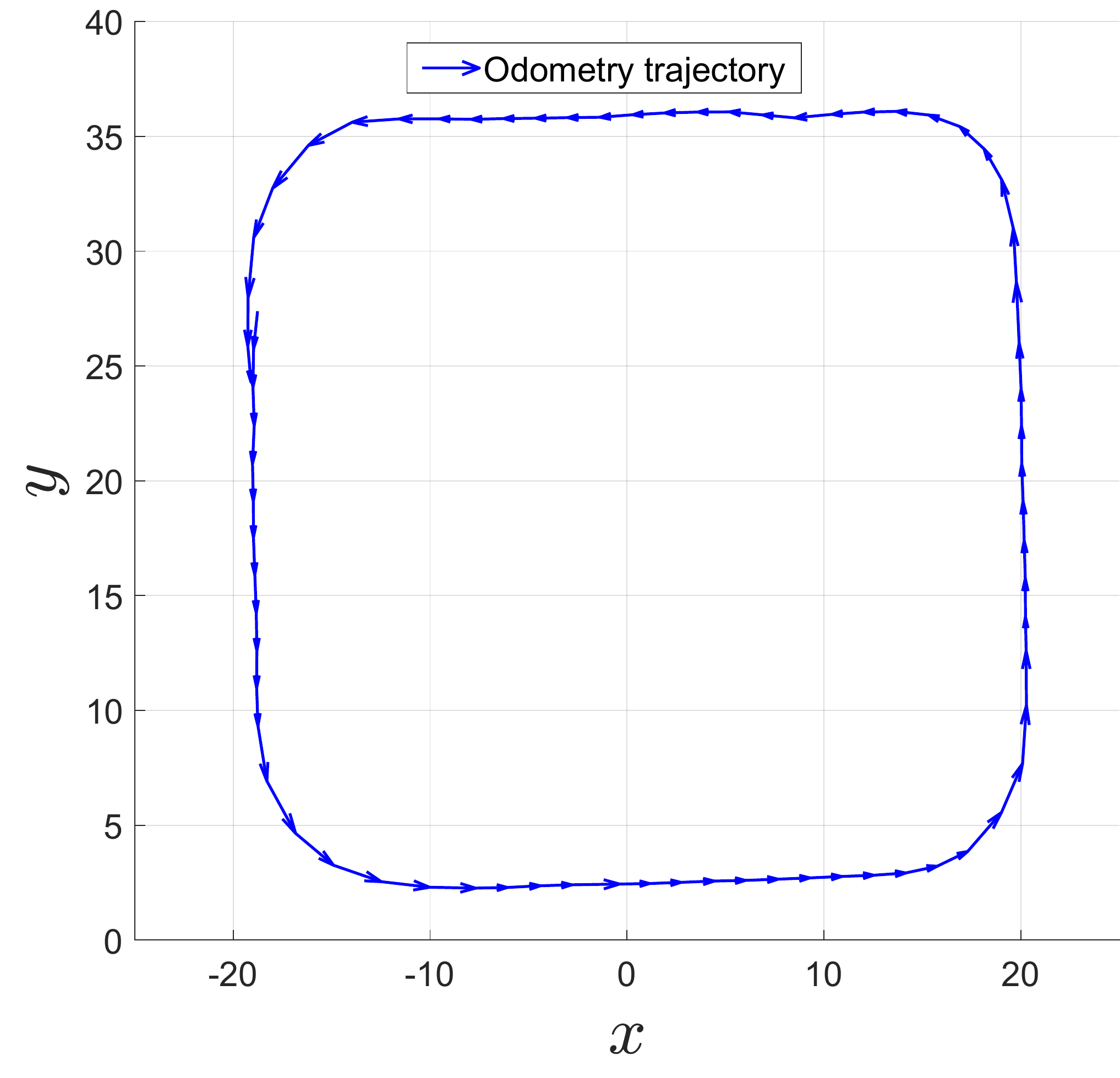}
		\caption{Perimeter monitoring}
		\label{fig:PM}
	\end{subfigure}%
	\begin{subfigure}[t]{0.25\textwidth}
		\centering
		\includegraphics[width=0.85\linewidth]{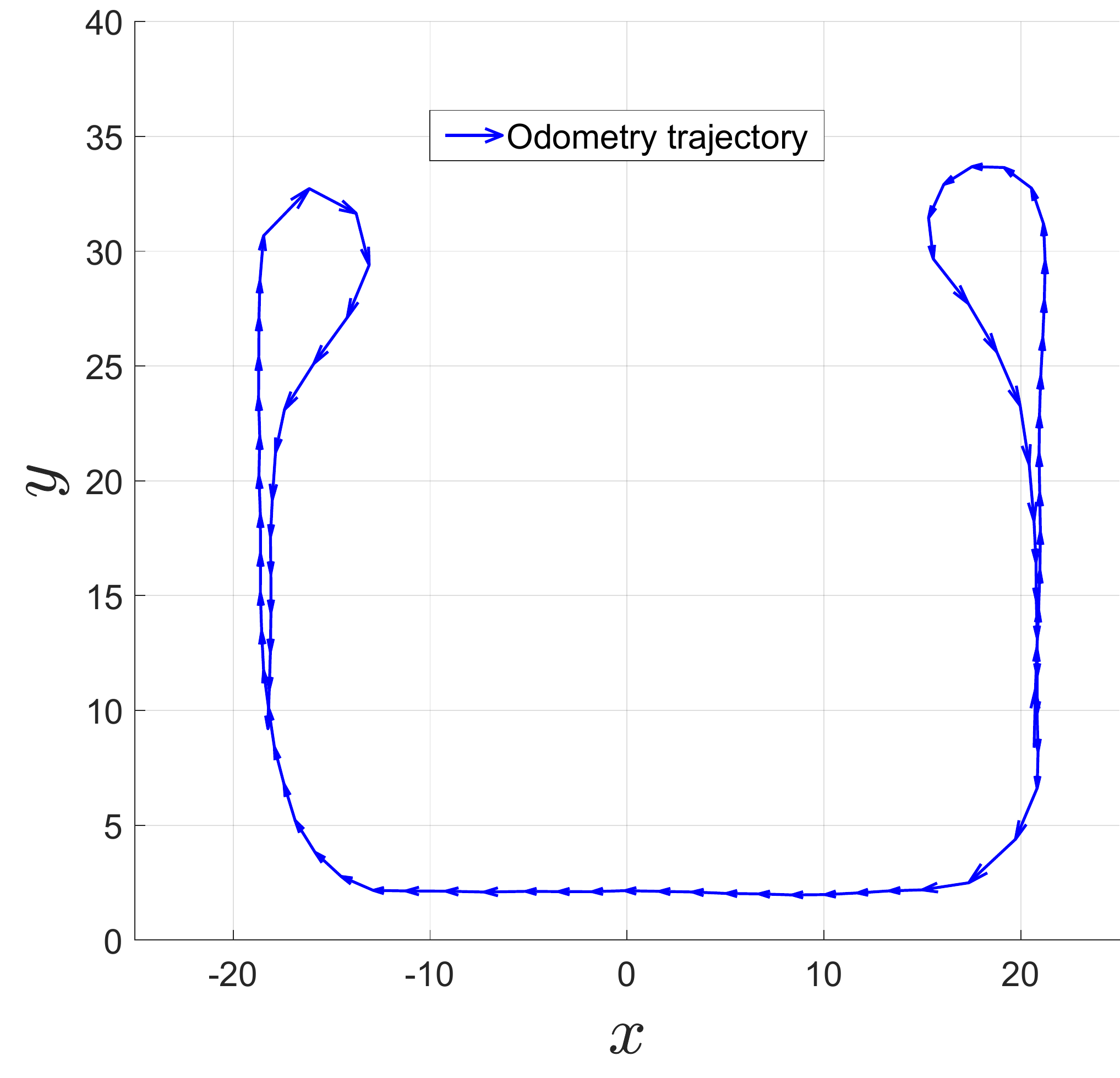}
		\caption{U-turn}
		\label{fig:U-turn}
	\end{subfigure}
	\caption{Odometry data}
	\label{fig:Position data}
\end{figure}

We use a dataset collected with the ``iCab'' vehicle (see Fig.\ref{fig:iCab}), while it moves in a closed environment, see Fig.\ref{fig:environment}. In this paper, we consider the internal information of the autonomous system that contains data related to the vehicle’s steering angle, rotor velocity, and power. We aim to detect unseen dynamics learned with the proposed method. Additionally, we want to select the best model. Two following situations are considered:

{\bf Perimeter monitoring task}. The vehicle follows a rectangular along the proposed environment, see Fig.\ref{fig:PM}. The information collected from the perimeter monitoring (PM) experience is employed as training data to learn our models.

\begin{figure}[H]
	\begin{subfigure}[t]{0.25\textwidth}
		\centering
		\includegraphics[width=0.85\linewidth]{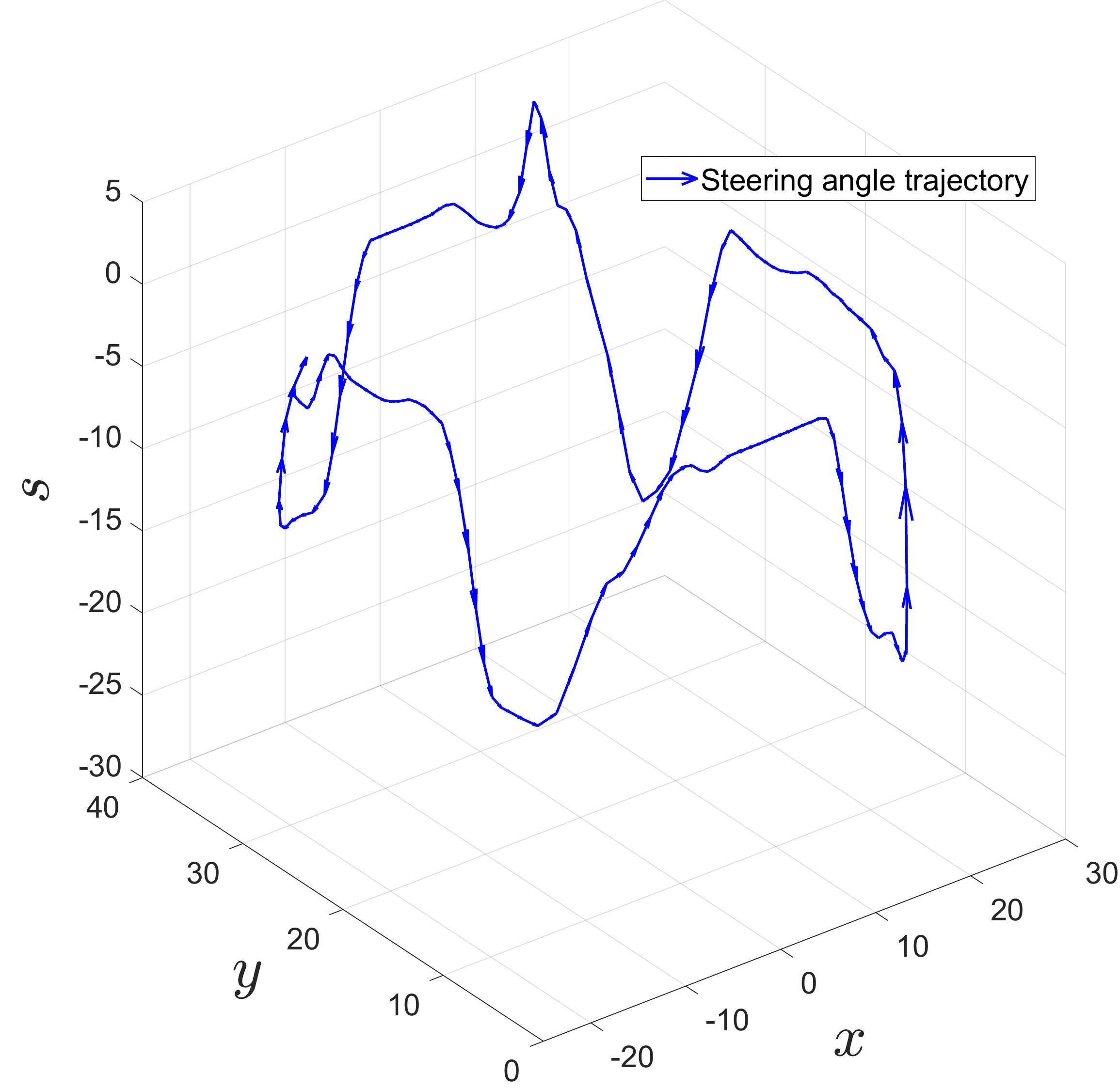}
		\caption{Steering w.r.t position data}
		\label{fig:SA}
	\end{subfigure}%
	\begin{subfigure}[t]{0.25\textwidth}
		\centering
		\includegraphics[width=0.85\linewidth]{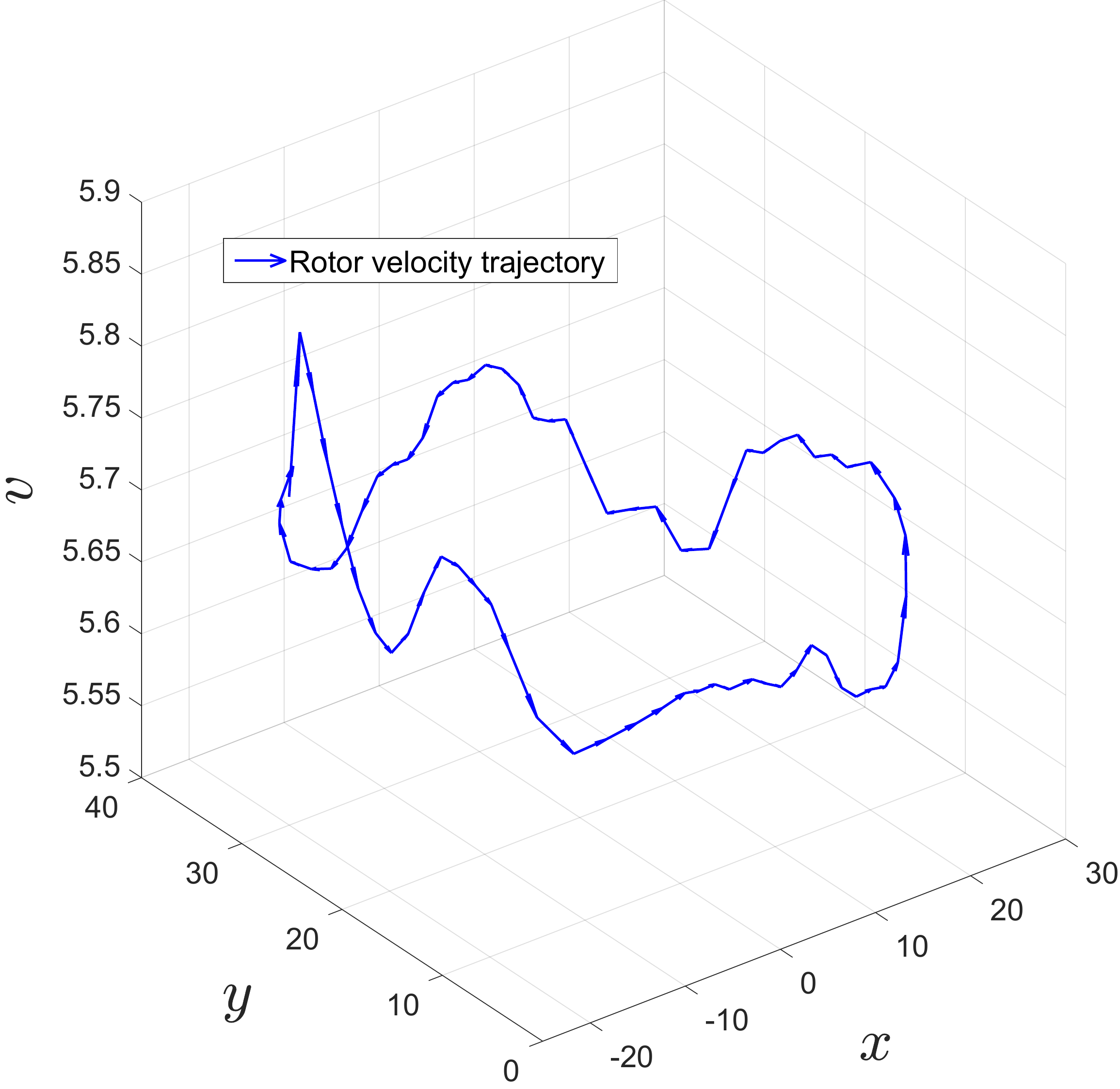}
		\caption{Rotor velocity w.r.t \\position data}
		\label{fig:V}
	\end{subfigure}
	\caption{Steering/Rotor data in a single lap of PM task}
	\label{fig:Control data}
\end{figure}

\begin{figure}[h]
\centering
 	\includegraphics[width = 0.85 \linewidth ]{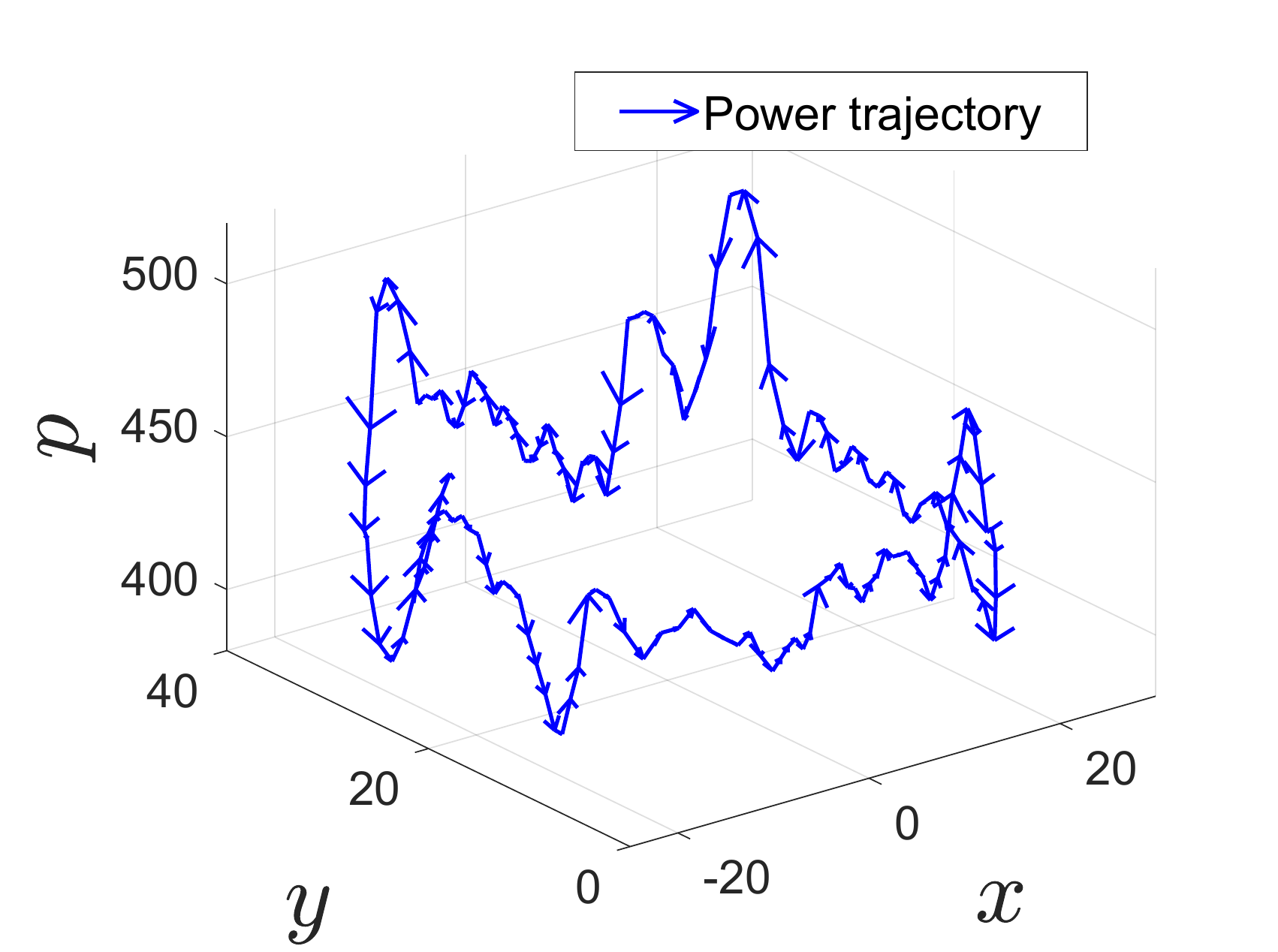}
\caption{Power w.r.t position data in a single lap of PM task}
\label{fig:P}
\end{figure}

The steering angle and velocity information are plotted with respect to position data in Fig.\ref{fig:SA} and  Fig.\ref{fig:V} respectively for a single lap of the experience. Additionally, The power information is shown in Fig.\ref{fig:P}.

{\bf U-turn task}. The vehicle performs the PM task until it encounters an obstacle. In such a case, the vehicle performs a U-turn manoeuvre, see Fig.\ref{fig:U-turn} and follows its rectangular path in the opposite direction. The information collected by this experiment is employed as testing data to detect anomalies in comparison to the PM task.

\subsection{Learning of DBNs}
In this work, we use different combinations of the collected information as a set of features from the vehicle. Such features are listed below:
\begin{itemize}
\item Steering angle, rotor Velocity and Power ($SVP$)
\item Steering angle and Power ($SP$)
\item Rotor Velocity and Power ($VP$)
\item Steering angle, rotor Velocity ($SV$)
\item Steering angle ($S$)
\item Rotor Velocity ($V$)
\item Power ($P$)
\end{itemize}

The features above mentioned represent the sensory data (cases) considered for building DBNs. Accordingly, for prediction and abnormality detection purposes we consider seven DBNs that follow the architecture shown in Fig. \ref{fig:DBN}. All proposed DBNs are trained based on the control data from the PM task, see Fig. \ref{fig:Control data} and Fig. \ref{fig:P}. For state prediction and detection of abnormalities, the control data produced by the U-turn experience is taken into consideration.  

\subsection{Abnormality detection and feature analysis}
\begin{figure*}[ht!]
\centering
 	(a)\includegraphics[width = 0.95 \linewidth ]{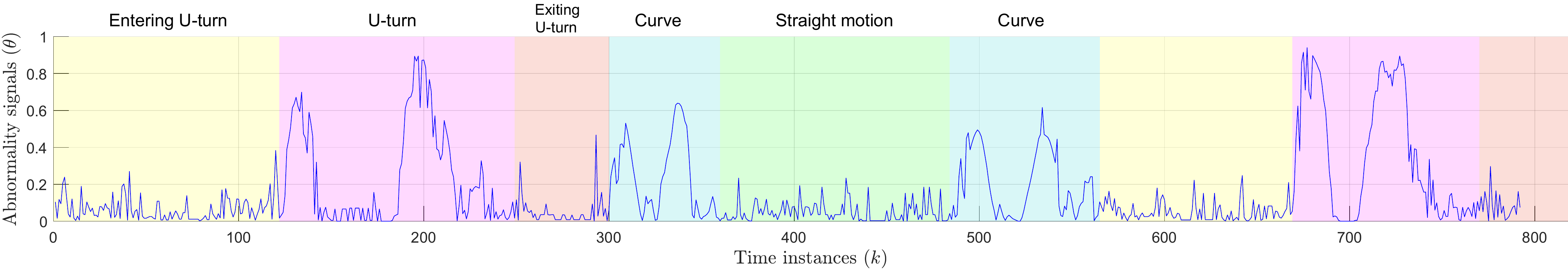}
 	(b)\includegraphics[width = 0.95 \linewidth ]{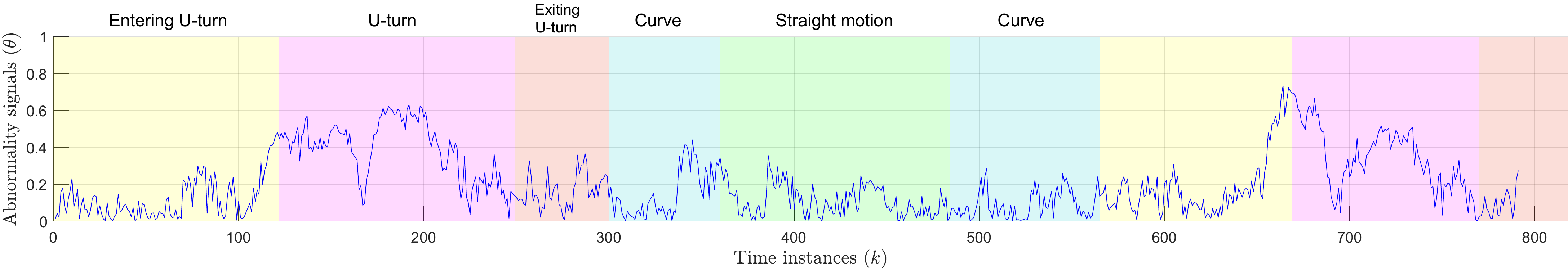}
 	(c)\includegraphics[width = 0.95 \linewidth ]{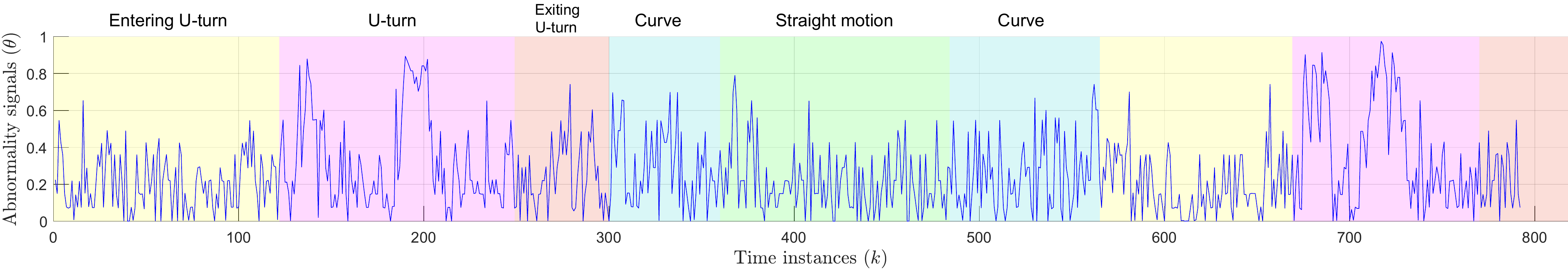}
\caption{Abnormality measurements for single variable features: (a) $S$, (b) $V$, and (c) $P$}
\label{fig:SVPsing}

 	(a)\includegraphics[width = 0.95 \linewidth ]{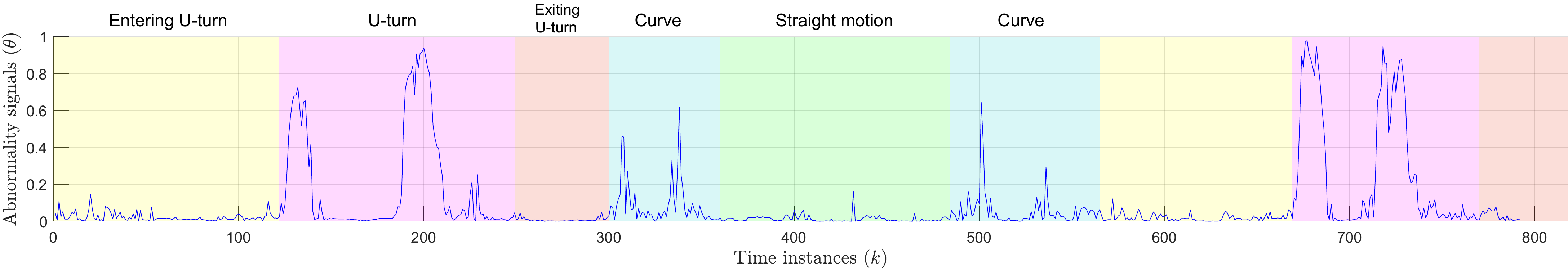}
 	(b)\includegraphics[width = 0.95 \linewidth ]{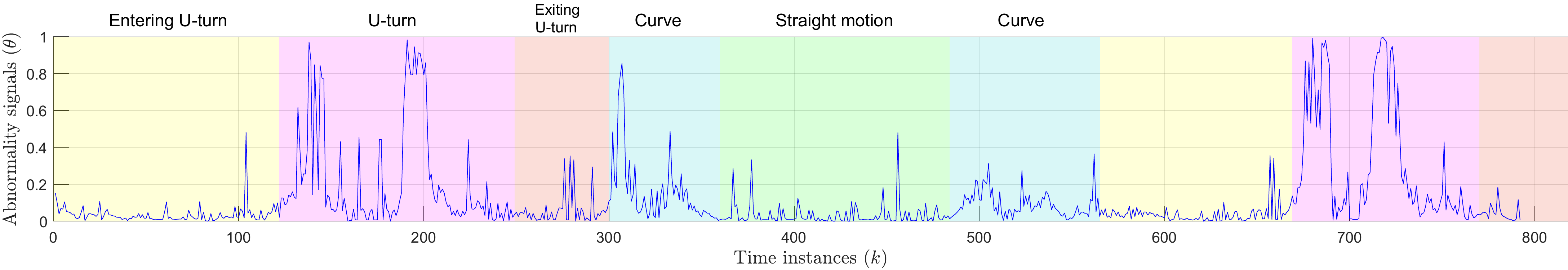}
 	(c)\includegraphics[width = 0.95 \linewidth ]{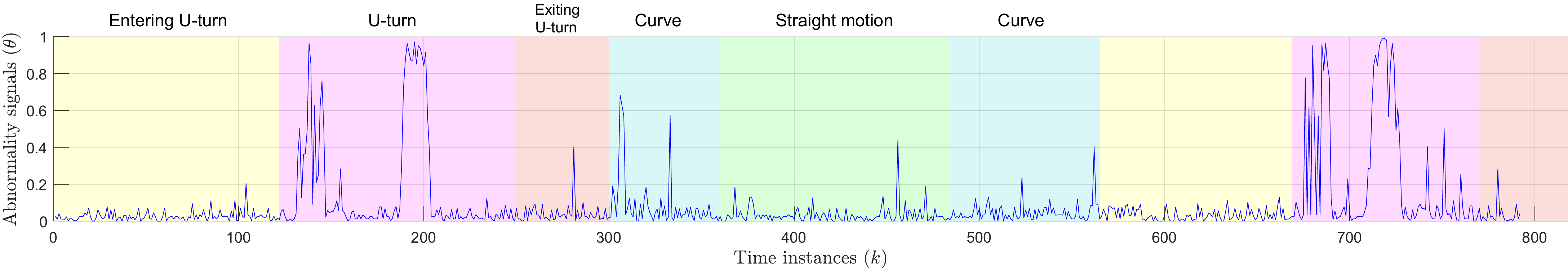}
 	(d)\includegraphics[width = 0.95 \linewidth ]{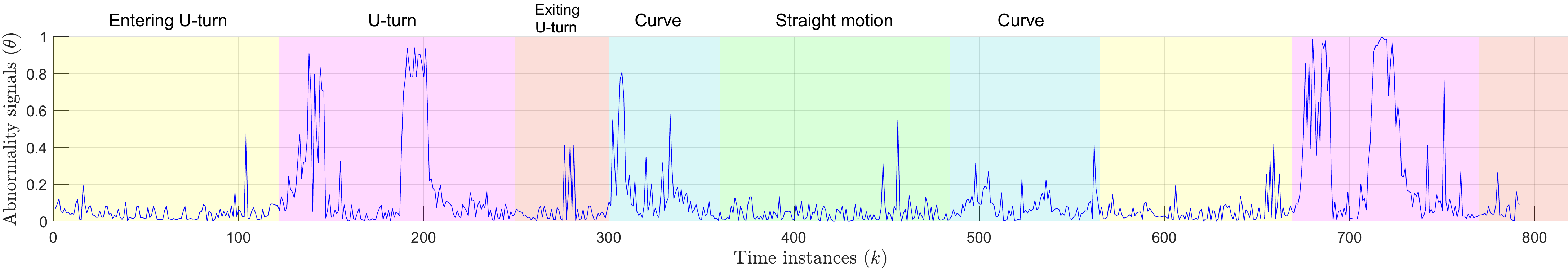}
\caption{Abnormality measurements for mixed variable features: (a) $SV$, (b) $SP$, (c) $VP$ and (d) $SVP$}
\label{fig:SVPcomb}
\end{figure*}

Based on the DBN feature-cases trained with the PM information, we perform prediction and detection of abnormalities with the features extracted from the U-turn observations. A manual ground truth (GT) of the vehicle's maneuvers is extracted for the U-turn experience. Fig. \ref{fig:SVPsing} and Fig. \ref{fig:SVPcomb} show the abnormality signals $\theta^m_k$, see equation \eqref{eq6}, through time for the different DBN feature-cases. Background colors of all plots are the same; they encode the GT of the vehicle's actions in the following way:
\begin{itemize}
\item Yellow: Entering U-turn (normal w.r.t PM) 
\item Violet: U-turn execution (abnormal w.r.t PM) 
\item Orange: Exiting U-turn (normal w.r.t PM)  
\item Blue: Inverse curve (abnormal w.r.t PM) 
\item Green: Straight motion (normal w.r.t PM)
\end{itemize}

\begin{figure}[t]
\centering
 	\includegraphics[width = 0.85 \linewidth ]{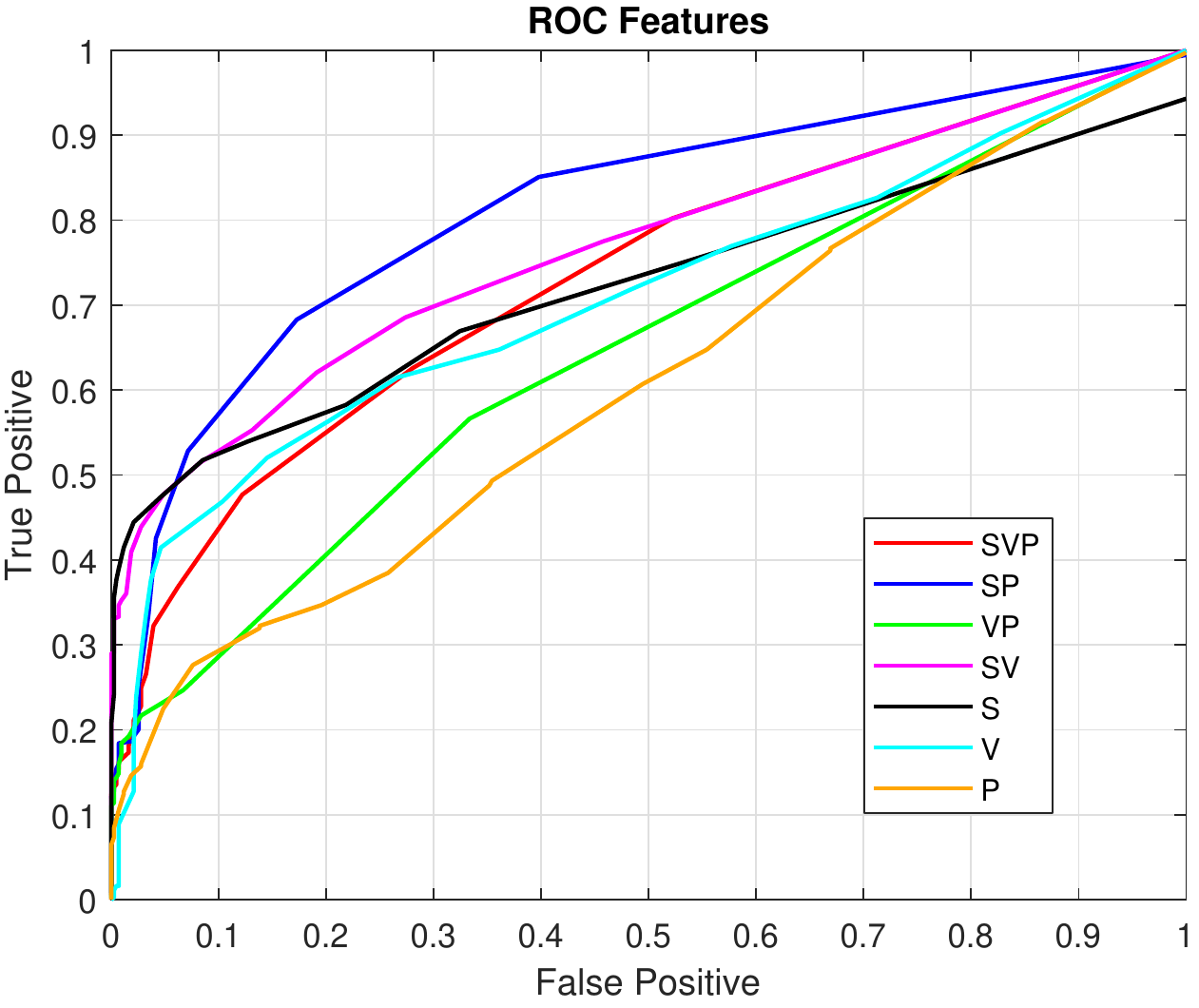}
\caption{ROC for different features}
\label{fig:roc}
\end{figure}

As can be seen, two new maneuvers w.r.t the PM are introduced in the U-Turn task. They are \textit{i)} The U-turn maneuver: which consists of a closed curve not experienced in the PM.    \textit{ii)} The inverse curve maneuver: that is present in the lower left/right parts of the trajectory shown in Fig.\ref{fig:U-turn}. Such a maneuver was learned in an anticlockwise sense in the PM. However, in the U-turn task, the vehicle performs such a curve also in clockwise direction. Based on the GT explained above, it is possible to build the ROC curve for each different feature-case. Fig. \ref{fig:roc} contains the ROC curves of the seven feature-cases. It is evident from the plot that $SP$ (curve in blue) performs better than the rest of the features. Table \ref{tab:table1} summarizes the performance comparison of different features based on the calculation of the two proposed precision measurements, $AUC$ of the ROC and $ACC$, see equation \eqref{eq9}. It can be seen that $SP$ presents the highest accuracy in both measurements w.r.t others features with values of $78.20\%$ and $76.11\%$ respectively. Such results suggest that the DBN trained with the $SP$ information provides the best recognition of abnormalities and the most accurate predictions of future instances.    

\pagebreak

\begin{table}[h]
    \centering
    \begin{tabular}{c|c|c|c|c|c|c|c}
         Feature &  $SVP$ & $SP$ & $VP$ & $SV$ & $S$ & $V$ & $P$ \\
         \hline
         $AUC$ (\%) & 72.99 & {\bf 78.20} & 64.27 & 76.76 & 71.44 & 70.71 & 60.44 \\
         $ACC$ (\%) & 69.40 & {\bf 76.11}  & 62.56 & 73.38 & 73.38 & 70.64 & 62.68 \\
    \end{tabular}
    \caption{Precision measurements}
    \label{tab:table1}
\end{table}

\section{Conclusions and future work}\label{conclusions and future work}
\label{sec:format}
This paper proposes and tests a method for selecting the most precise DBN when predicting abnormalities in real scenarios where multiple sensory data is analyzed. Our method is evaluated with real data taken from a moving vehicle performing some tasks in a closed environment.

Results suggest that the proposed method recognizes the features from a set of sensory data that facilitates the recognition of previously unseen manoeuvres, i.e., abnormal situations. The proposed method can be useful for selecting relevant features when dealing with a large amount of features in networking operations.      

As future work, the proposed approach can be adapted to identify and characterize cross-correlations between a selected coupled DBNs to provide complementary information for prediction and classification purposes. 

\vfill\pagebreak

\bibliographystyle{plain}
\bibliography{WFIOT}

\begin{thebibliography}{10}

\bibitem{atzori2010internet}
Luigi Atzori, Antonio Iera, and Giacomo Morabito.
\newblock The internet of things: A survey.
\newblock {\em Computer networks}, 54(15):2787--2805, 2010.

\bibitem{baydoun2018learning}
M~Baydoun, D~Campo, V~Sanguineti, L~Marcenaro, A~Cavallaro, and C~Regazzoni.
\newblock Learning switching models for abnormality detection for autonomous
  driving.
\newblock In {\em 2018 21st International Conference on Information Fusion
  (FUSION)}, pages 2606--2613. IEEE, 2018.

\bibitem{baydoun2018multi}
Mohamad Baydoun, Mahdyar Ravanbakhsh, Damian Campo, Pablo Marin, David Martin,
  Lucio Marcenaro, Andrea Cavallaro, and Carlo~S Regazzoni.
\newblock A multi-perspective approach to anomaly detection for self-aware
  embodied agents.
\newblock {\em arXiv preprint arXiv:1803.06579}, 2018.

\bibitem{Bhattacharyya1943}
A.~Bhattacharyya.
\newblock {On a measure of divergence between two statistical populations
  defined by their probability distributions}.
\newblock {\em Bulletin of the Calcutta Mathematical Society}, 35:99--109,
  1943.

\bibitem{camara2017self}
Javier C{\'a}mara, Kirstie~L Bellman, Jeffrey~O Kephart, Marco Autili, Nelly
  Bencomo, Ada Diaconescu, Holger Giese, Sebastian G{\"o}tz, Paola Inverardi,
  Samuel Kounev, et~al.
\newblock Self-aware computing systems: Related concepts and research areas.
\newblock In {\em Self-Aware Computing Systems}, pages 17--49. Springer, 2017.

\bibitem{chatila2017toward}
Raja Chatila, Erwan Renaudo, Mihai Andries, Ricardo-Omar Chavez-Garcia, Pierre
  Luce-Vayrac, Rapha{\"e}l Gottstein, Rachid Alami, Aur{\'e}lie Clodic, Sandra
  Devin, Beno{\^\i}t Girard, et~al.
\newblock Toward self-aware robots.
\newblock {\em Frontiers in Robotics and AI}, 5:88, 2017.

\bibitem{Fritzke:1994:GNG:2998687.2998765}
Bernd Fritzke.
\newblock A growing neural gas network learns topologies.
\newblock In {\em Proceedings of the 7th International Conference on Neural
  Information Processing Systems}, NIPS'94, pages 625--632, Cambridge, MA, USA,
  1994. MIT Press.

\bibitem{green2015internet}
Harriet Green.
\newblock The internet of things in the cognitive era: Realizing the future and
  full potential of connected devices.
\newblock {\em ed: IBM Watson IoT}, 2015.

\bibitem{han2017tdoa}
Seul-Ki Han, Won-Sang Ra, and Jin~Bae Park.
\newblock Tdoa/fdoa based target tracking with imperfect position and velocity
  data of distributed moving sensors.
\newblock {\em International Journal of Control, Automation and Systems},
  15(3):1155--1166, 2017.

\bibitem{1160055}
J.~O. Kephart and D.~M. Chess.
\newblock The vision of autonomic computing.
\newblock {\em Computer}, 36(1):41--50, Jan 2003.

\bibitem{Lourenzutti2014}
Rodolfo Lourenzutti and Renato~A. Krohling.
\newblock The hellinger distance in multicriteria decision making: An
  illustration to the topsis and todim methods.
\newblock {\em Expert Syst. Appl.}, 41(9):4414--4421, July 2014.

\bibitem{rao2018feature}
Haidi Rao, Xianzhang Shi, Ahoussou~Kouassi Rodrigue, Juanjuan Feng, Yingchun
  Xia, Mohamed Elhoseny, Xiaohui Yuan, and Lichuan Gu.
\newblock Feature selection based on artificial bee colony and gradient
  boosting decision tree.
\newblock {\em Applied Soft Computing}, 2018.

\bibitem{ravanbakhsh2018learning}
Mahdyar Ravanbakhsh, Mohamad Baydoun, Damian Campo, Pablo Marin, David Martin,
  Lucio Marcenaro, and Carlo~S Regazzoni.
\newblock Learning multi-modal self-awareness models for autonomous vehicles
  from human driving.
\newblock {\em arXiv preprint arXiv:1806.02609}, 2018.

\bibitem{wu2014cognitive}
Qihui Wu, Guoru Ding, Yuhua Xu, Shuo Feng, Zhiyong Du, Jinlong Wang, and Keping
  Long.
\newblock Cognitive internet of things: a new paradigm beyond connection.
\newblock {\em IEEE Internet of Things Journal}, 1(2):129--143, 2014.

\end{thebibliography}

\end{document}